\documentclass[letterpaper, 10 pt, conference]{ieeeconf}  % Comment this line out if you need a4paper
\IEEEoverridecommandlockouts                              % This command is only needed if 
\usepackage{caption}
\usepackage{subcaption}
\usepackage{graphicx}
\graphicspath{{Figures/}}

\usepackage{amssymb}
\usepackage{amsmath} % Required for some math elements 
\usepackage{mathtools}
\usepackage{dsfont}
\usepackage{bbold}

\usepackage{algorithmic}
\usepackage{array}
\usepackage{lipsum}
\usepackage{hyperref}
\usepackage{color}
\usepackage{dsfont}

\definecolor{red}{RGB}{187,0,0}
\definecolor{blue}{RGB}{0, 0,180}

\newcommand{\Exp}[1]{\mathbb{E}[ #1]}
\usepackage{nth}
\usepackage{cite}

\title{\LARGE \bf
Human-in-the-Loop Framework for Real-World  Deep Reinforcement Learning Algorithms: Salient Features, Challenges and Trade-offs
}

\title{\LARGE \bf
A Systematic Approach to Design Real-World Human-in-the-Loop Deep Reinforcement Learning: Salient Features, Challenges and Trade-offs
}

\author{Jalal Arabneydi$^{1}$, Saiful Islam$^2$, Srijita Das$^2$, Sai Krishna Gottipati$^3$, William Duguay$^3$, Cloderic Mars$^3$, \\
Matthew E. Taylor$^2$, Matthew Guzdial$^2$, Antoine Fagette$^4$ and Younes Zerouali$^{1}$% <-this % stops a space
\thanks{$^{1}$ Jalal Arabneydi and Younes Zerouali are with JACOBB, 10555, avenue de Bois-de-Boulogne,
Montreal, Quebec, Canada, postal code: H4N 1L4,
        {\tt\small jalal.arabneydi@mail.mcgill.ca} and {\tt\small younes.zerouali@jacobb.ai}}%
\thanks{$^{2}$ Saiful Islam, Srijita Das,  Matthew Taylor and Matthew Guzdial  are with the Department of Computer Science, University of Alberta and Amii (Alberta Machine Intelligence Institute), Athabasca Hall, 9119 - 116 St NW, Edmonton, Alberta, Canada, postal code:
T6G 2E8,
        {\tt\small mdsaifu1@ualberta.ca,srijita1@ualberta.ca, matthew.e.taylor@ualberta.ca, guzdial@ualberta.ca} }%
\thanks{$^{3}$ Sai Krishna Gottipati, William Duguay and Cloderic Mars are with AI Redefined (AIR), 400 McGill St number 300, Montreal, Quebec, Canada, postal code: H2Y 2G1,
        {\tt\small sai@ai-r.com},{\tt\small william@ai-r.com} and {\tt\small cloderic@ai-r.com}}%      
\thanks{$^{4}$ Antoine Fagette is Research and Technology Lead in AI at Thales, 6650 Rue Saint-Urbain Bureau 350, Montréal, Quebec, Canada, postal code: H2S 3G9
        {\tt\small antoine.fagette.e@thalesdigital.io}}%   
}

\begin{document}

\date{\today} % Date for the report

\maketitle
\thispagestyle{empty}
\pagestyle{empty}

%%%%%%%%%%%%%%%%%%%%%%%%%%%%%%%%%%%%%%%%%%%%%%%%%%%%%%%%%%%%%%%%%%%%%%%%%%%%%%%%
\begin{abstract}
With the growing popularity of deep reinforcement learning (DRL),  human-in-the-loop (HITL) approach has the potential to revolutionize the way we approach decision-making problems and create new opportunities for human-AI collaboration. In this article, we introduce a novel multi-layered hierarchical HITL DRL algorithm that comprises three types of learning: self learning, imitation learning and transfer learning. In addition, we consider three forms of human inputs: reward, action and demonstration. Furthermore,  we discuss main challenges, trade-offs and advantages of HITL in solving complex problems and how  human information can be integrated in the AI solution systematically. To verify our technical results, we present a real-world unmanned aerial vehicles (UAV) problem wherein a number of enemy drones attack a restricted area. The objective is to design a scalable HITL DRL algorithm for ally drones to neutralize the enemy drones before they reach the area. To this end, we first implement our solution using an award-winning open-source HITL software called Cogment. We then
demonstrate several interesting results such as (a) HITL leads to faster training and higher performance, (b)  advice acts as a guiding direction for gradient methods and lowers variance, and (c) the amount of advice should neither be too large nor too small to avoid over-training and under-training. Finally, we illustrate the role of human-AI cooperation in solving two real-world complex scenarios, i.e., overloaded and decoy attacks.

\end{abstract}

\section{Introduction}

Human-in-the-loop (HITL) deep reinforcement learning (DRL) is a hybrid approach that combines the strengths of human intelligence and machine learning algorithms to improve the performance of RL agents in real-world applications. The approach involves incorporating human feedback into the reinforcement learning process, providing guidance and direction to the DRL agent~\cite{mnih2015human,silver2016mastering,arulkumaran2017deep}.

Applications of HITL DRL are wide-ranging and include robotics, autonomous systems, gaming, and more. In robotics, HITL DRL has been used to train robots to perform tasks that require fine-grained manipulation and interaction with the environment~\cite{levine2018learning}. In autonomous systems, the approach has been used to improve the performance of autonomous vehicles, enabling them to make safer and more efficient decisions~\cite{Cai2020DRL_autonomous,WU2022}. In gaming, HITL DRL has been used to develop game AI that can adapt to human opponents, leading to more challenging and engaging game play~\cite{Christiano2017_deepmind}.

One of the key challenges in HITL DRL is designing an effective mechanism for incorporating human feedback into the reinforcement learning process. This requires considering factors such as the type of feedback, the frequency and timing of feedback, and the format of feedback~\cite{taylor2014reinforcement}. Additionally, the design of human-AI interaction interfaces is also an important consideration, as it impacts the ease and efficiency of providing feedback~\cite{weitekamp2020interaction}.

Recent advancements in this field include methods for incorporating human feedback into the reinforcement learning process in a scalable and efficient manner. For example, some studies have proposed using inverse reinforcement learning to estimate human preferences~\cite{kim2022application}, while others have proposed using human feedback to fine-tune  DRL agents~\cite{warnell2018deep}. Other studies have focused on the design of human-AI interaction interfaces, exploring ways to make the feedback process more intuitive and efficient~\cite{abedin2022designing}.

The outline of this paper is as follows. In Section~\ref{sec:prob}, we present a problem formulation of HITL multi-agent RL and discuss its main challenges. In Section~\ref{sec: proposed_sol}, we propose a multi-layered hierarchical HITL DRL algorithm, where each layer can be constructed in three steps. We also mention a few salient features and trade-offs of HITL solutions. In Section~\ref{sec:UAV}, we describe an UAV problem and in Section~\ref{sec:UAV_sol}, we provide our proposed solution. In Section~\ref{sec:result}, we illustrate our technical results with numerical simulations. In Section~\ref{sec:conclusion}, we conclude the paper.

\subsection{Notations and conventions}
$\mathbb{N}$ and $\mathbb{R}$ denote the sets of natural and real numbers, respectively. Given any $k \in \mathbb{N}$, $\mathbb{N}_k$ denotes the set $\{1, 2, \ldots,k\}$, and given any $a \leq b \in \mathbb{N}$, $x_{a:b}$ denotes the vector $(x_a,\ldots,x_b)$. For any orientation $\phi \notin [-\pi,\pi]$, $\pm 2\pi$ is added iteratively until $\phi \in [-\pi,\pi]$. In addition, $\|\boldsymbol \cdot \|_p$ denotes $L_p$ norm and $\mathds{1}(\boldsymbol \cdot)$ is indicator function. For any vector position $p \in \mathbb{R}^2$, $p^x$ and $p^y$ refer to the $x$ and $y$ positions, respectively. Notation $\lor$ represents logical ``or".

\section{Problem formulation}\label{sec:prob}
For the purpose of generalization, we consider a multi-agent RL formulation, where single-agent RL is a special case. In particular, suppose we have a dynamic stochastic environment wherein $n \in \mathbb{N}$ agents wish to collaborate together to accomplish a common task. Let $x^e_t \in \mathbb{R}^{d^e_x}$ and $w^e_t \in \mathbb{R}^{d^e_w}$ denote the state and uncertainty of the environment at time $t \in \mathbb{N}$, respectively, where $d^e_x,d^e_w \in \mathbb{N}$. Let also $x^i_t \in \mathbb{R}^{d_x}$, $u^i_t \in \mathbb{R}^{d_u}$ and $w^i_t \in \mathbb{R}^{d_w}$ be the state, action and noise of agent $i \in \mathbb{N}_n$ at time $t$, respectively, where $d_x, d_u, d_w \in \mathbb{N}$. To ease the display, we use bold letters to denote joint variables, i.e., $\mathbf x_t := (x^e_t, x^1_t, \ldots,x^n_t) \in \mathcal{X}$, $\mathbf u_t := (u^1_t, \ldots,u^n_t) \in \mathcal{U}$ and $\mathbf w_t := (w^e_t, w^1_t, \ldots,w^n_t) \in \mathcal{W}$, where $\mathcal{X}$, $\mathcal{U}$ and $\mathcal{W}$ are feasible spaces of joint state, action and noise, respectively.

At any time $t$, suppose that joint state evolves as follows:
\begin{equation}
\mathbf x_{t+1} = f(\mathbf x_t, \mathbf u_t, \mathbf w_t),
\end{equation}
where function $f$ is called dynamics. Denote by $o^i_t \in \mathbb{R}^{d_o}$ and $v^i_t \in \mathbb{R}^{d_v}$ the observation and measurement noise of agent $i$ at time $t$, respectively, where $d_o, d_v \in \mathbb{N}$. Define $\mathbf o_t := (o^1_t,\ldots,o^n_t)$ and $\mathbf v_t := (v^1_t,\ldots,v^n_t)$ such that
\begin{equation}
\mathbf o_t = z(\mathbf x_t, \mathbf v_t ),
\end{equation}
where function $z$ is called observation function. It is assumed that noise processes in the dynamics and observation are mutually independent across time horizon.   Define strategy $\mathbf g := \{g^1,\ldots,g^n\}$, where $g^i, i \in \mathbb{N}_n,$ denotes the control law of agent $i$, i.e.,
\begin{equation}
u^i_t = g^i(o^i_{1:t}, u^i_{1:t-1}).
\end{equation}
It is to be noted that the information structure is decentralised meaning that every agent has access to a different set of information. Define a team reward function as follows:
\begin{equation}\label{eq:reward}
\mathbb{E}(\frac{1}{n}\sum_{t=1}^\infty  \sum_{i=1}^n  \gamma^{t-1} r^i(\mathbf x_t, \mathbf u_t)),
\end{equation}
where   $r^i(\mathbf x_t, \mathbf u_t) \in \mathbb{R}$ is the reward of agent $i$ and
$\gamma \in (0,1)$ is a discount factor. 

\indent \textit{HITL problem:}
We are interested to develop a scalable multi-agent RL algorithm  to maximize \eqref{eq:reward} with the help of human input (information).

\textbf{Main challenges:}
Below, we mention a few computational and conceptual challenges in solving HITL problem.
\begin{enumerate}

\item Suppose  for now that there is no human in the loop. In such a case, solving the above problem is difficult because it requires a very large set of samples due to the continuous state and action spaces, imperfect observations, and non-linear stochastic unknown dynamics and reward functions; to mention only a few. In addition, when information structure is decentralized,  every agent has a different perspective about the global system, making it conceptually challenging to establish cooperation among agents.  Furthermore, computational complexity in space is exponential with respect to the number of agents, i.e., it is not scalable. 

\item On the other hand, having human in the loop can alleviate some of the above challenges; however, it  comes at the cost of dealing with another set of challenges. For example, it is not clear what, where, when, and how much human feedback must be provided. Moreover, HITL is time-consuming and expensive, in general. Plus, human feedback may be inconsistent due to many factors such as mood, motivation and personal biases. In addition, it may be harmful and misleading as well as difficult to be utilized at scale.

\end{enumerate}

\section{Proposed solution}\label{sec: proposed_sol}
To simplify the exposition and focus attention on human interactions, we assume that states are estimated by an state estimator and can be treated as Markov processes. With a slight abuse of notation, we use $\mathbf x$ for the joint state estimate. Therefore, one can write the following Bellman equation:
\begin{align} \label{eq:bellman}
&V(\mathbf x) = \max_{\mathbf u \in \mathcal{U}}(\frac{1}{n}  \sum_{i=1}^n r^i(\mathbf x, \mathbf u) + \mathbb{E}_{\mathbf w}(V(f(\mathbf x, \mathbf u, \mathbf w)))),\\
&Q(\mathbf x, \mathbf u) :=\frac{1}{n}  \sum_{i=1}^n r^i(\mathbf x, \mathbf u) + \mathbb{E}_{\mathbf w}(V(f(\mathbf x, \mathbf u, \mathbf w))),\\
&\mathbf g^* := \text{argmax}_{\mathbf u \in \mathcal{U}} \hspace{.1cm} Q(\boldsymbol \cdot, \mathbf u).
\end{align}

Since solving~\eqref{eq:bellman} is difficult, we use human input to facilitate the process of finding a good solution. To this end, define strategy $\mathbf g_H := \{g^1_H,\ldots,g^n_H\}$, where $g^i_H, i \in \mathbb{N}_n,$ denotes the control law of agent $i$ under human influence. As an example, one can consider a case when human input is directly encoded into action space, e.g.,
\begin{equation}
u^i_t = \phi^i_H u^i_{H,t} + (1-\phi^i_H) u^i_{A,t},
\end{equation}
where $\phi^i_H \in \{0,1\}$ is an indicator function choosing between human action $u^i_{H,t}$ and AI action $u^i_{A,t}$. Similarly, one can define value function $V_H$ and Q-function $Q_H$. The ideal goal is to converge to the optimal solution in a more efficient way  using human help, i.e.,
\begin{equation}
\lim_{k \rightarrow \infty }  V^k_H =  V^\ast, \lim_{k \rightarrow \infty }  Q^k_H =  Q^\ast, \lim_{k \rightarrow \infty } \mathbf g^k_H = \mathbf g^\ast, 
\end{equation}
where $k \in \mathbb{N}$ denotes the index of iterations. To be practical, we often look for a good (approximate) solution rather than the optimal one in real-world scenarios.

\begin{figure}[t!]
\center
\includegraphics[scale=.33, trim = {.9cm 2.5cm 1cm 0cm}, clip]{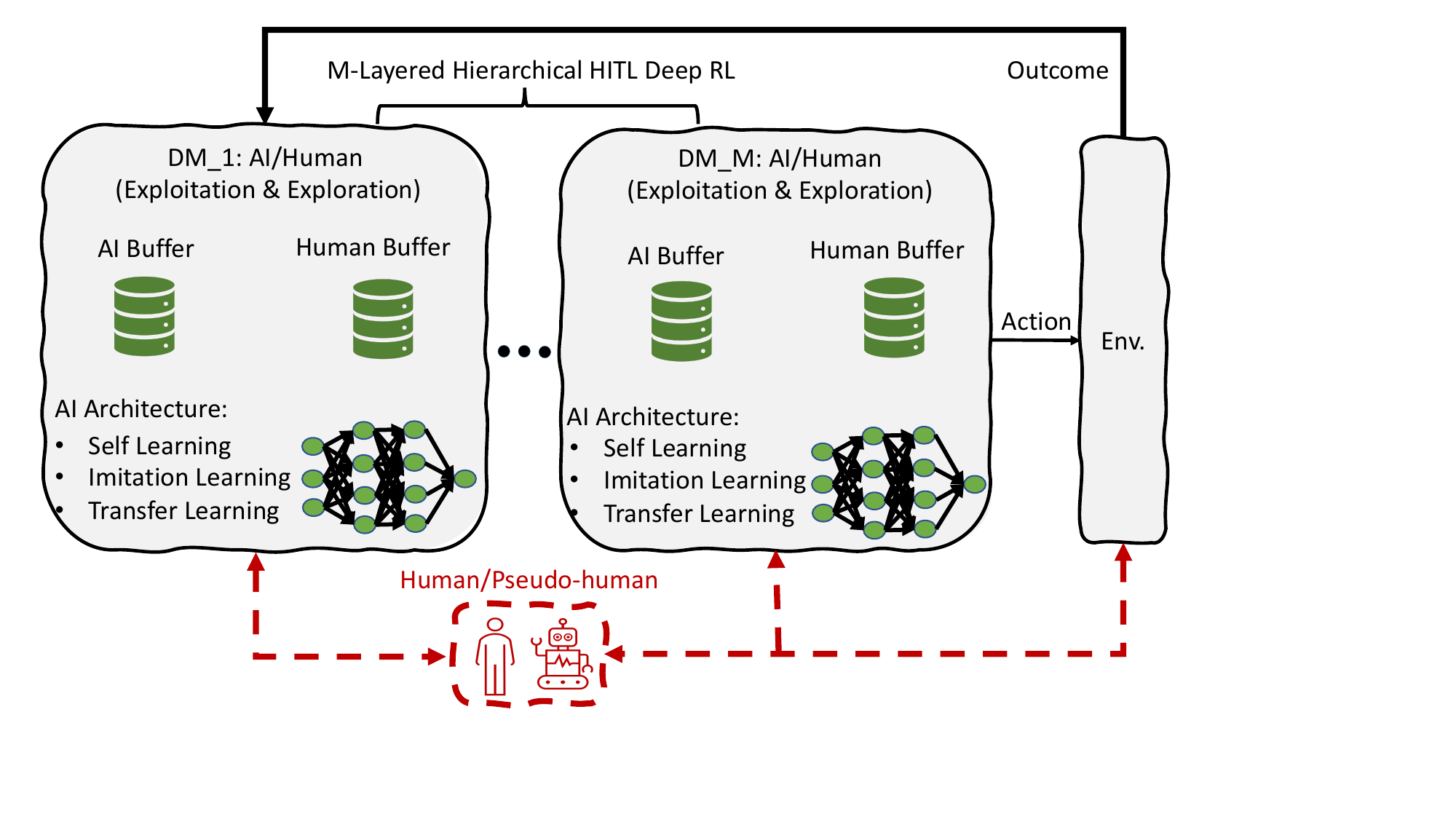}
\caption{General structure of a multi-layered hierarchical HITL deep reinforcement leaning algorithm. }\label{fig:hierarchical}
\end{figure}

In general, one has to take into account three fundamental questions when it comes to training AI with HITL:
\begin{itemize}
\item \textbf{self learning:} how to learn independently for cases in which no human input is available,
\item \textbf{imitation learning:} how to imitate human strategy for cases in which human data is available,
\item \textbf{transfer learning:} how to transfer learned intuitions and skills to another applications. 
\end{itemize}
In addition, to cope with the scalability challenge, we often use hierarchical approaches. Figure~\ref{fig:hierarchical} depicts a general layout and components of a multi-layered hierarchical HITL deep RL, where human data can be used directly or indirectly (via a pseudo-human, e.g., demonstration network).
In what follows, we use an actor-critic formulation~\cite{Sutton2018introduction} for one layer of the hierarchical structure in Figure~\ref{fig:hierarchical} to demonstrate  a HITL deep RL algorithm in three steps.  Similar formulation can be used for other RL algorithms as well as other layers.  Suppose critic and actor functions are modelled by deep feed-forward neural networks. We now follow three steps.

\noindent\textbf{Step 1: Data generation (exploration and exploitation).} Given a diminishing probability $\epsilon_k  \in [0,1]$, one can perform $\epsilon-$greedy policy with a distinction  that at the exploration phase, human can also provide his/her action with an independent probability $\phi_{H,k} \in [0,1]$, i.e.,
\begin{equation}
\mathbf u_k = \begin{cases}
\mathbf g_H(\mathbf x_k) + \mathbf w^{\text{en}}_k, & \text{w.p. }(1-\phi_{H,k})\times \epsilon_k, \\ 
\mathbf u_{H,k}, & \text{w.p. }\phi_{H,k} \times \epsilon_k,\\
\mathbf g_H(\mathbf x_k), &\text{w.p. } (1-\epsilon_k),\\
\end{cases}
\end{equation}
where $\mathbf w^{\text{en}}_k$ is the exploration noise and $\mathbf u_{H,k}$ is the action given by human at iteration $k \in \mathbb{N}$. It is to be noted that probability $\phi_{H,k}$ can be influenced by human, meaning that human can decide when to provide guiding action $\mathbf u_{H,k}$.

\noindent\textbf{Step 2: Sampling from Human and AI Buffers.} At every iteration $k$, the outcome is stored in either human buffer $\mathcal{B}_H$ or AI buffer $\mathcal{B}_A$,  depending on whose action generated the outcome. In particular, the outcome includes current state, current action, immediate reward, next state and whether or not the task is done. 

Let $\mathbf x^{\text{train}}$, $\mathbf u^{\text{train}}$, $\mathbf r^{\text{train}}$, $\mathbf x^{\text{train}}_+$ and $\mathbf{done}^{\text{train}}$ be a batch sample drawn from $\mathcal{B}_H$ and $\mathcal{B}_A$. The ratio of samples from $\mathcal{B}_H$ over those from $\mathcal{B}_A$ indicates the amount of human advice in the training phase. For example, one can define \emph{advice probability} by specifying the probability according to which samples are drawn from $\mathcal{B}_H$.

\noindent\textbf{Step 3: Learning.} At iteration $k \in \mathbb{N}$, let $\mathbf r^{\text{train}}_{H,k}$ denote an additional reward provided by human. Through this reward, human can guide the RL algorithm towards a desired behaviour. Define
\begin{equation}
\mathbf Q^k_{\text{target}}:= \mathbf r^{\text{train}}_k+ \mathbf r^{\text{train}}_{H,k} +  \gamma(1-\mathbf{done}^{\text{train}}_k)  Q^k_H(\mathbf x^{\text{train}}_{+,k}, \mathbf g^k_H(\mathbf x^{\text{train}}_{+,k})).
\end{equation}
In addition, define the following critic's loss function
\begin{equation}
L^k_{Q}  \hspace{-.1cm}:=  \hspace{-.1cm}\frac{1}{n_B} \hspace{-.1cm}\left( \hspace{-.05cm} \| Q^k_H(\mathbf x^{\text{train}}_k \hspace{-.05cm}, \hspace{-.05cm}\mathbf u^{\text{train}}_k)  \hspace{-.05cm}- \hspace{-.05cm}\mathbf Q^k_{\text{target}}\|^2_2  \hspace{-.1cm}+  \hspace{-.1cm} \lambda^k_Q \| Q^k_H\hspace{-.05cm}(\mathbf x^{\text{train}}_k \hspace{-.05cm}, \hspace{-.05cm} \mathbf u^{\text{train}}_k) \|^2_2 \right)
\end{equation}
where $n_B$ is the batch size and $\lambda^k_Q \in [0,\infty)$ is regularization coefficient at iteration $k$. The critic can be updated as:
\begin{equation}
 Q^{k+1}_H =
  Q^k_H - \alpha^k_Q \nabla L^k_Q,
\end{equation}
where $\alpha^k_Q \in (0,\infty)$ is the step size. Now, define actor's loss function such that
\begin{align}
L^k_g &:=\frac{1}{n_B} \big(\beta_k \big[-  \sum_{j=1}^{n_B} Q^k_H(x^{\text{train}}_k(j),u^{\text{train}}_k(j)) \big]
\nonumber \\
&\quad + (1-\beta_k) \big[\|\mathbf g^k_H(\mathbf x^{\text{train}}_k) - \mathbf u^{\text{train}}_k \|^2_2\big] + \lambda^k_g \|\mathbf g^k_H(\mathbf x^{\text{train}}_k) \|^2_2 \big),
\end{align}
where  $\beta_k \in [0,1]$ is a deciding factor for the trade-off between self learning and imitation learning, and $\lambda^k_g \in [0,\infty)$ is a regularization coefficient. For example, $\beta_k = 1$ favours self learning and $\beta_k =0$ favours imitation learning. Hence, the actor function can be updated as 
\begin{equation}
 \mathbf g^{k+1}_H =
  \mathbf g^k_H - \alpha^k_g \nabla L^k_g,
\end{equation}
where $\alpha^k_g \in (0,\infty)$ is the step size. It is to be noted that one can convert human demonstrations into proper rewards, actions and policies in order to use the above HITL setup.

\textbf{Salient features:} In what follows, we present a few features of HITL solution.

\begin{itemize}
\item \textbf{Hybrid decision-making \& higher performance}: HITL combines the strengths of both machine learning and human decision-making, leading to better performance and adaptability.

\item \textbf{Faster training \& less computation}: HITL alleviates sample inefficiency problem of real-world RL algorithms, saving a significant amount of computational resources, e.g., time and memory.

\item \textbf{Safe \& robust}: HITL systems are safer and more reliable to changing circumstances and environments, as they can learn and respond to new information provided by human. In other words, humans can provide feedback and intervene in the learning process to prevent undesirable or harmful behaviour.

\item \textbf{Intuitive \& transparent}: HITL is an intuitive and transparent approach in order to take advantage of human skill sets and develop human-level AI.

\end{itemize}

\textbf{Key trade-offs:} When using HITL, one has to take into account several key trade-offs such as the following.
\begin{itemize}
\item \textbf{Too much vs. too little advice}: Quantity of information matters because too much information may lead to over-training while too little may lead to under-training.
\item \textbf{Computation vs. labour cost}: To generate AI's data, one has to spend lots of money on computational resources such as data storage and GPU. On the the hand, human data can be expensive too, specially when it comes to hiring human experts.
\item \textbf{Good vs. bad advice}: Quality of information matters because good quality information helps AI (i.e., positive effect) while bad one misleads it (i.e., negative effect).
\item \textbf{Simple vs. complex task}: Efficiency of HITL and transferring skill sets from human to AI depends on the the difficulty level of tasks. For example, simple tasks are often preferred for learning macro-behaviours (i.e., basics)  while complex tasks are more suited for micro behaviours (i.e., nuances).
\end{itemize}

\section{Unmanned Aerial Vehicle (UAV) Example}\label{sec:UAV}
\begin{figure}[t!]
\center
\includegraphics[scale=.075]{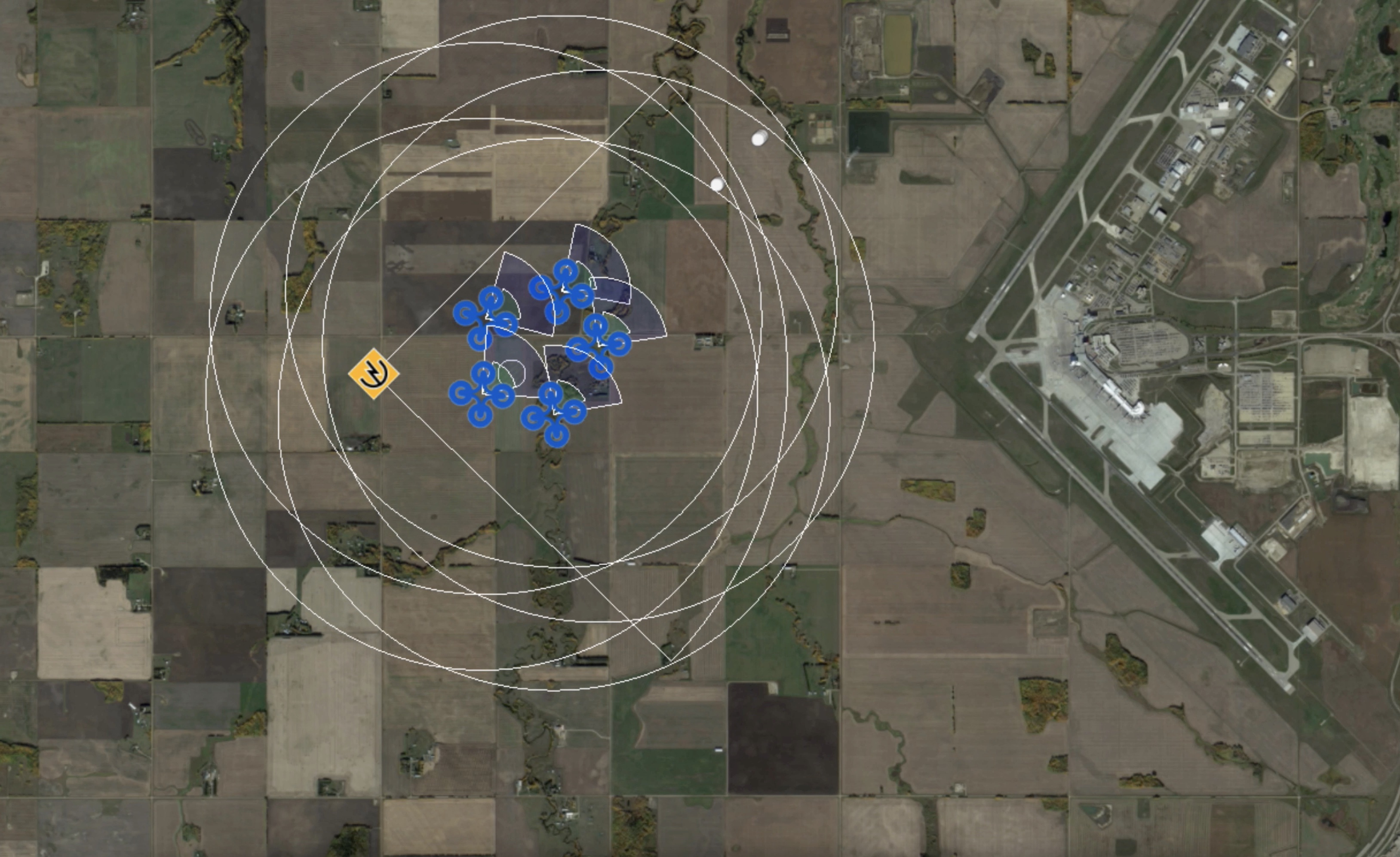}
\caption{A snapshot of UAV example simulated in Cogment~\cite{Cogment2021}, where yellow square is ground radar, blue quad-rotors are ally drones and white circles are enemy drones. The field of view of radars are displayed by white lines.}\label{fig:screen}
\end{figure}

In this section, we illustrate our proposed solution via a simplified real-world UAV example.
Consider a scenario in which a few \textit{enemy} drones attack a restricted area and a few \textit{ally} drones wish to neutralize them  before they reach the area, as displayed in Figure~\ref{fig:screen}. In what follows, we describe a mathematical formulation of the above scenario. Let $p_{\text{RA}} \in \mathbb{R}^2$ and $r_{\text{RA}} \in [0,\infty)$ be the center and radius of a circle representing the restricted area, respectively.

\subsection{Ally system}
The ally system consists of one ground control station (GCS) and several ground radars (GR) and aerial drones. 

\subsubsection{Ally ground control station (GCS)}
There is one GCS located at position $p^{\text{GC}}_t \in \mathbb{R}^2$ at time step $t \in \mathbb{N}$.  In addition, let $r_{\text{GC}} \in (0, \infty)$ denote the radius of a circle representing the operating range of the GCS.

\subsubsection{Ally ground radar (GR)}
Consider $n_{\text{GR}} \in \mathbb{N}$  ground radars whose jobs are to gather information about the ally and enemy drones. Denote by $p^{i,\text{GR}}_t \in \mathbb{R}^2$ the position and by $\phi^{i,\text{GR}}_t \in [-\pi, \pi]$ the heading angle (orientation) of the $i$-th GR at time $t \in \mathbb{N}$, $i \in \mathbb{N}_{n_{\text{GR}}}$. Let $u^{i,\text{GR}}_t \in [-1, 1]$ denote the action of the $i$-th GR, where $u^{i,\text{GR}}_t$ is the ratio of angular speed with respect to its maximum value. In particular, $u^{i,\text{GR}}_t$ controls the rotation of the $i$-th GR as follows:
\begin{equation}
\phi^{i,\text{GR}}_{t+1} = \phi^{i,\text{GR}}_t +  v^{\text{max}}_{\text{GR}} \times u^{i,\text{GR}}_t,
\end{equation}
where $v^{\text{max}}_{\text{GR}} \in [0, \infty)$ is the maximum angular speed. Denote by $\rho_{\text{GR}} \in [0, 2 \pi]$ the field of view of the GR and by $r_{\text{GR}} \in [0, \infty)$ the radius of a circle representing its sensing range. 

\subsubsection{Ally drone}\label{Sec:ally_state}
Let $n_{\text{AD}} \in \mathbb{N}$ be the number of ally drones. The state of ally drone $i \in \mathbb{N}_{ n_{\text{AD}}}$ is as follows:
\begin{enumerate}
\item position $p^i_t \in \mathbb{R}^2$,

\item heading angle (orientation) $\phi^i_t \in [-\pi, \pi]$,

\item relative orientation of the electro-optic (EO) sensor, i.e.,  $\phi^{i,\text{EO}}_t \in [- \phi^{\text{max}}_{\text{EO}}, \phi^{\text{max}}_{\text{EO}}]$ (mounted on drone) with respect to the drone's heading, where $\phi^{\text{max}}_{\text{EO}} \in [0, \frac{300}{180}\pi]$,

\item whether or not the drone is functional, i.e.,  $f^i_t \in \{0,1\}$, where $0$ means the drone is not functional,

\item whether or not controlled by GCS, i.e.,  $g^i_t \in \{0,1\}$, where $0$ means the drone is not controlled by GCS,

\item position of GCS controlling drone $i$, i.e., $p^{i,\text{GC}}_t \in \mathbb{R}^2$,

\item whether or not the radar is enabled, i.e., $\text{radar-enabled}^i_t \in \{0,1\}$, where $0$ means the radar (mounted on the drone) is off,

\item whether or not electro-magnetic pulse (EMP) has been used so far at least once, i.e., $\text{emp-used}^i_t \in \{0, 1\}$, where $0$ means that EMP has never been used. In particular, there is an EMP-auto-destruction probability $pr_{\text{EMPD}} \in [0,1]$ according to which the drone may destroy itself upon using EMP, i.e.,
\begin{equation}
f^i_{t+1} = \begin{cases}
0, & f^i_t = 0,\\
1, & f^i_t = 1, \text{emp-used}^i_t = 0, \\
B, & f^i_t = 1, \text{emp-used}^i_t = 1, 
\end{cases}
\end{equation}
where $B \in \{0,1\}$ is a Bernoulli random variable with success probability of $1 - pr_{\text{EMPD}}$.
\end{enumerate}
Denote by $\rho_{\text{EO}} \in [0,2\pi]$ the field of view of the EO sensor and by $r_{\text{EO}} \in [0, \infty]$ the radius of a circle representing its sensing range. In addition, let $\rho_{\text{AD}} \in [0, 2\pi]$ be the field of view of the radar and $r_{\text{AD}} \in [0, \infty)$ be the radius of a circle representing its sensing range. In addition, the action set of drone $i$ at time $t$ is described below.
\begin{enumerate}
\item control signal for the movement angle $u^{i,\text{MA}}_t \in [-\pi, \pi]$,

\item seed-ratio of the drone, i.e., $u^{i,\text{SR}}_t \in [0,1]$ such that
\begin{equation}
p^i_{t+1} = p^i_t +   v^{\text{max}}_{\text{SR}} \times u^{i,\text{SR}}_t  \begin{bmatrix}
\cos(u^{i,\text{MA}}_t)\\
\sin(u^{i,\text{MA}}_t)
\end{bmatrix},
\end{equation}
where $ v^{\text{max}}_{\text{SR}} \in (0, \infty)$ is the maximum speed of drone,

\item angular speed-ratio $u^{i,\text{heading}}_t \in [-1, 1]$ such that
\begin{equation}
\phi^i_{t+1} = \phi^i_t +  v^{\text{max}}_{\text{AS}} \times  u^{i,\text{heading}}_t,
\end{equation}
where $v^{\text{max}}_{\text{AS}} \in (0, \infty)$ is the maximum angular speed,

\item angular speed-ratio of EO sensor $u^{i,\text{EO}}_t \in [-1, 1]$ s.t.
\begin{equation}
\phi^{i,\text{EO}}_{t+1} = \begin{cases}
-\phi^{\text{max}}_{\text{EO}},  & \phi^{i,\text{EO}}_t \leq -\phi^{\text{max}}_{\text{EO}}, \\
+\phi^{\text{max}}_{\text{EO}}, & \phi^{i,\text{EO}}_t \geq \phi^{\text{max}}_{\text{EO}}, \\
 \phi^{i,\text{EO}}_t +  v^{\text{max}}_{\text{ASEO}} \times u^{i,\text{EO}}_t, & \text{otherwise},\\
 \end{cases}
\end{equation}
where $v^{\text{max}}_{\text{ASEO}} \in (0, \infty)$ is the maximum angular speed of EO sensor.

\item turn off/on the drone's radar $u^{i,\text{ER}}_t \in \{0,1\}$, where $0$ refers to turning the radar off, i.e.,
\begin{equation}
\text{radar-enabled}^i_{t+1} = u^{i,\text{ER}}_t,
\end{equation}

\item turn off/on the EMP $u^{i,\text{EMP}}_t \in \{0,1\}$, where $0$ refers to turning the EMP off, i.e.,
\begin{equation}
\text{emp-used}^i_{t+1} = u^{i,\text{EMP}}_t + \text{emp-used}^i_t (1 -u^{i,\text{EMP}}_t),
\end{equation}

\item turn off/on jamming, i.e., $u^{i,\text{EJ}}_t \in \{0,1\}$, where $0$ refers to turning the jamming off,

\item turn off/on GPS spoofing, i.e., $u^{i,\text{GPSS}}_t \in \{0,1\}$, where $0$ refers to turning the spoofing off,

\item turn off/on hacking, i.e., $u^{i,\text{EH}}_t \in \{0,1\}$, where $0$ refers to turning the hacking off.
\end{enumerate} 

\subsection{Enemy system}
The enemy system consists of $n_{\text{EGCS}}  \in \mathbb{N}$ number of GCSs and $n_{\text{ED}}  \in \mathbb{N}$ number of drones.

\subsubsection{Enemy ground control station (GCS)}

Let $p^i_{\text{EGCS}}(t) \in \mathbb{R}^2$ denote the position of the $i$-th enemy's GCS at time $t$, where $i \in \mathbb{N}_{n_{\text{EGCS}}}$. Also, let $r_{\text{EGCS}} \in [0, \infty]$ denote the radius of a circle representing its operating range.

\subsubsection{Enemy drone}\label{Sec:enemy_state}

The state of enemy drone $i \in \mathbb{N}_{n_{\text{ED}}}$ at time $t$ is given as follows:
\begin{enumerate}
\item position $p^{i,\text{ED}}_t \in \mathbb{R}^2$,

\item payload $\ell^{i,\text{ED}}_t \in \{0,1,2, 3\}$, where $0$ means ``safe'', $1$ ``unknown'', $2$ ``moderate'', and $3$ ``dangerous",

\item whether or not drone $i$ is controlled by a GCS, i.e. $g^{i,\text{ED}}_t \in \{0,1\}$, where $0$ means the drone is not controlled by the GCS,

\item position of the GCS controlling drone $i$,  $p^{i,\text{EGCS}}_t \in \mathbb{R}^2$,

\item whether or not the drone is functional, i.e., $f^{i,\text{ED}}_t \in \{0,1\}$, where $0$ means it is not functional; in particular,
\begin{equation}
f^{i,\text{ED}}_{t+1} \hspace{-.1cm}=\hspace{-.1cm} \begin{cases}
0, & \exists j \in \mathbb{N}_{n_{\text{AD}}} \hspace{.1cm} s.t.  \hspace{.1cm} f^j_t\times  u^j_t =1\\
 &   \qquad \text{and} \hspace{.1cm}  \|p^{i,\text{ED}}_t - p^j_t\|_2 \leq r_{\text{N}},\\
0, &  \|p^{i,\text{ED}}_t - p^{i,\text{EGCS}}_t\|_2 > r_{\text{EGCS}}, g^{i,\text{ED}}_t\hspace{-.1cm} =\hspace{-.1cm}1,\\
f^{i,\text{ED}}_t, & \text{otherwise},
\end{cases}
\end{equation}
where $r_{\text{N}} \in [0,\infty)$ is the radius of a circle representing the neutralization range of ally drones and $u^j_t \in \{0,1\}$ is defined as whether or not the ally drone $j \in \mathbb{N}_{n_{\text{AD}}}$ is enabled to neutralize enemy drone $i$, i.e.,
\begin{equation}\label{eq:nut_action}
u^j_t \hspace{-.1cm}:=\hspace{-.1cm} u^{j,\text{EMP}}_t \lor u^{j,\text{GPSS}}_t \lor \big(g^{i,\text{ED}}_t  u^{j,\text{EJ}}_t\big) \lor \big(g^{i,\text{ED}}_t u^{j,\text{EH}}_t\big).
\end{equation}

\end{enumerate}

\subsection{Observation}\label{Sec:observation}
In this manuscript, it is assumed that the states of ally devices are observed perfectly. However, those of enemy devices are observed with measurement noises using  various sensors, meaning that we often have more than one estimated position for one target.
\subsubsection{Measurement noise}
Let $p^{\text{TA}}_t \in \mathbb{R}^2$ be the location of a target to be measured and $p_t \in \mathbb{R}^2$ be the location of a sensor (i.e., ground radar, drone radar, EO sensor, and radio-frequency sensor). Given the relative position between the target and the sensor, i.e. $\Delta p_t := p^{\text{TA}}_t -p_t $, define two measurement noises:
\begin{equation}
\Delta p^{\text{D}}_t:= \frac{\Delta p_t}{\|\Delta p_t \|_2} \times d_{\text{DE}} \times \text{uniform}(-0.5, 0.5),
\end{equation}
and 
\begin{equation}
\Delta p^{\text{A}}_t:= \Delta p_t^\perp \times d_{\text{AE}} \times \text{uniform}(-0.5, 0.5),
\end{equation}
where $d_{\text{DE}} \in [0, \infty)$ is the linear-error percentage and $d_{\text{AE}} \in [0, \infty)$ is the angular-error percentage of the sensor. It is to be noted that $\text{uniform}(a,b)$ denotes uniform probability distribution function on interval $[a,b]$ and $\Delta p^\perp_t :=(\Delta p^y_t, -\Delta p^x_t)$ is a vector perpendicular to $\Delta p_t$, i.e., $\Delta p^\intercal_t \Delta p^\perp_t =0$. 
Define stochastic function $h: \mathbb{R}^2 \rightarrow \mathbb{R}^2$ as follows:
\begin{equation}
h(\Delta p_t):= \Delta p^{\text{D}}_t + \Delta p^{\text{A}}_t.
\end{equation}
Hence, the location of the target $p^{\text{TA}}_t$ is observed as:
\begin{equation}
o^\text{TA}_t := p^\text{TA}_t + h(\Delta p_t).
\end{equation}

\subsubsection{Detection by ground radar (GR)}
Let $p^{\text{TA}}_t \in \mathbb{R}^2$ be the location of a target at time $t$ and $p^{\text{GR}}_t \in \mathbb{R}^2$ be the location of a GR. Define the relative position as follows:
\begin{equation}\label{eq:relative_pos}
\Delta p_t := p^{\text{TA}}_t - p^{\text{GR}}_t,
\end{equation}
and relative orientation as:
\begin{equation}\label{eq:relative_orien}
\Delta \phi_t := \arctan(\Delta p^y_t, \Delta p^x_t).
\end{equation}
The GR detects the target with probability $pr_{\text{GR}} \in [0,1]$ if the target is within its sensing range and field of view, i.e.,
\begin{equation}\label{eq:GR_obs}
o^{\text{GR}}_t \hspace{-.1cm}=\hspace{-.1cm} \begin{cases}
\hat o^{\text{GR}}_t, & \|\Delta p_t \|_2 \leq r_{\text{GR}},\Delta \phi_t\hspace{-.1cm} \in\hspace{-.1cm} [\phi_{\text{GR}}(t) \pm \frac{\rho_{\text{GR}}}{2}],\\
\text{blank}, & \text{otherwise},
\end{cases}
\end{equation}
where
\begin{equation}\label{eq:GR_obs_prob}
\hat o^{\text{GR}}_t := \begin{cases}
p^{\text{TA}}_t + h(\Delta p_t), & \text{with probability } pr_{\text{GR}},\\
\text{blank}, &   \text{otherwise}.
\end{cases}
\end{equation}

\subsubsection{Detection by drone radar}
Drone radar detects an object in a similar fashion that the GR does in~\eqref{eq:GR_obs} and~\eqref{eq:GR_obs_prob}. In particular,  define appropriate relative position and orientation similar to~\eqref{eq:relative_pos} and~\eqref{eq:relative_orien}. Then, the drone radar detects a target with probability $pr_{\text{DR}} \in [0,1]$ if the target is within its sensing range of radius $r_{\text{AD}}$ and field  of view $\rho_{\text{AD}}$ radians.

%
%\subsubsection{Detection by drone radar}
%Drone radar detects an object in a similar fashion that the GR does. Let $p_{\text{T}}(t) \in \mathbb{R}^2$ be the location of a target  at time $t$ and $p(t) \in \mathbb{R}^2$ be the location of the drone. Define the relative position as follows:
%\begin{equation}
%\Delta p(t) := p_{\text{T}}(t) - p(t),
%\end{equation}
%and relative orientation as:
%\begin{equation}
%\Delta \phi(t) := \arctan(\Delta p^y(t), \Delta p^x(t)).
%\end{equation}
%The drone radar detects the target with probability $pr_{\text{DR}} \in [0,1]$ if the target is within its sensing range and field of view, i.e.,
%\begin{equation}
%o_{\text{DR}}(t)\hspace{-.1cm} := \hspace{-.1cm}\begin{cases}
%\hat o_{\text{DR}}(t), & \|\Delta p(t) \|_2 \leq r_{\text{AD}}, \Delta \phi(t) \hspace{-.1cm}\in\hspace{-.1cm} [\phi(t) \pm \frac{\rho_{\text{AD}}}{2}],\\
%\text{blank}, & \text{otherwise},
%\end{cases}
%\end{equation}
%where
%\begin{equation}
%\hat o_{\text{DR}}(t) := \begin{cases}
%p_{\text{T}}(t) + h(\Delta p(t)), & \text{with probability } pr_{\text{DR}},\\
%\text{blank}, &   \text{otherwise}.
%\end{cases}
%\end{equation}

\subsubsection{Detection by EO sensor}
EO sensor detects an object in a similar fashion  in~\eqref{eq:GR_obs} and~\eqref{eq:GR_obs_prob} with a distinction that $EO$ sensor's orientation also depends on the the drone's orientation. More precisely, the actual orientation is $\phi_t + \phi_{\text{EO}}$. Therefore, define appropriate relative position and orientation similar to~\eqref{eq:relative_pos} and~\eqref{eq:relative_orien}. Then, the EO sensor detects a target with probability $pr_{\text{EO}} \in [0,1]$ if the target is within its sensing range of radius $ r_{\text{EO}}$  and field of view of $\rho_{\text{EO}}$ radians. In addition, let $r_{\text{EO-payload}} \in [0,\infty)$. If $o^{\text{EO}}_t \neq \text{blank}$ and $\|\Delta p_t\|_2 \leq r_{\text{EO-payload}}$, then EO sensor can also detect the payload of the enemy drone.

%\subsubsection{Detection by EO sensor}
%EO sensor detects an object in a similar fashion that the GR does with a difference that $EO$ sensor's orientation also depends on the the drone's orientation. In particular, let $p_{\text{T}}(t) \in \mathbb{R}^2$ be the location of a target  at time $t$ and $p(t) \in \mathbb{R}^2$ and $\phi(t) \in [-\pi, \pi]$ be the location and orientation of the drone carrying the EO sensor, respectively. Define the relative position as follows:
%\begin{equation}
%\Delta p(t) := p_{\text{T}}(t) - p(t),
%\end{equation}
%and relative orientation as:
%\begin{equation}
%\Delta \phi(t) := \arctan(\Delta p^y(t), \Delta p^x(t)).
%\end{equation}
%The EO sensor detects the target with probability $pr_{\text{EO}} \in [0,1]$ if the target is within its sensing range and field of view, i.e.,
%\begin{equation}
%o_{\text{EO}}(t)\hspace{-.1cm} :=\hspace{-.1cm} \begin{cases}
%\hat o_{\text{EO}}(t), & \|\Delta p(t) \|_2 \leq r_{\text{EO}}, \Delta \phi(t) \in [\phi(t) + \phi_{\text{EO}}(t)\pm \frac{\rho_{\text{EO}}}{2}],\\
%\text{blank}, & \text{otherwise},
%\end{cases}
%\end{equation}
%where
%\begin{equation}
%\hat o_{\text{EO}}(t) := \begin{cases}
%p_{\text{T}}(t) + h(\Delta p(t)), & \text{with probability } pr_{\text{EO}},\\
%\text{blank}, &   \text{otherwise}.
%\end{cases}
%\end{equation}
%Let $r_{\text{EO-payload}} \in [0,\infty)$. If $o_{\text{EO}}(t) \neq \text{blank}$ and $\|\Delta p(t)\|_2 \leq r_{\text{EO-payload}}$, then EO sensor can also detect the payload of the enemy drone.
%

\subsubsection{Detection by radio-frequency (RF) sensor}
RF sensor detects an enemy in a similar fashion  in~\eqref{eq:GR_obs} and~\eqref{eq:GR_obs_prob} with a distinction that the enemy drone must be controlled by GCS, i.e., $g^{\text{ED}}_t = 1$. In particular, define appropriate relative position and orientation similar to~\eqref{eq:relative_pos} and~\eqref{eq:relative_orien}. Then, RF sensor detects an enemy with probability $pr_{\text{RF}} \in [0,1]$ if the target is within its sensing range of radius $r_{\text{RF}}$.

%
%\subsubsection{Detection by radio-frequency (RF) sensor}
%RF sensor is able to detect an enemy in the case when an enemy drone is controlled by GCS, i.e., $g_{\text{ED}} = 1$. In particular, let
%$r_{\text{RF}} \in [0, \infty)$ be the radius of a circle representing the sensing range of the RF sensor. Moreover, let 
%$p_{\text{T}}(t) \in \mathbb{R}^2$ be the location of a target  at time $t$ and $p(t) \in \mathbb{R}^2$  be the location of the drone carrying the RF sensor, respectively. Define the relative position as follows:
%\begin{equation}
%\Delta p(t) := p_{\text{T}}(t) - p(t),
%\end{equation}
%and relative orientation as:
%\begin{equation}
%\Delta \phi(t) := \arctan(\Delta p^y(t), \Delta p^x(t)).
%\end{equation}
%The RF sensor detects the target with probability $pr_{\text{RF}} \in [0,1]$ if the target is within its sensing range, i.e.,
%\begin{equation}
%o_{\text{RF}}(t) := \begin{cases}
%\hat o_{\text{RF}}(t), & \|\Delta p(t) \|_2 \leq r_{\text{RF}}, g_{\text{ED}} = 1,\\
%\text{blank}, & \text{otherwise},
%\end{cases}
%\end{equation}
%where
%\begin{equation}
%\hat o_{\text{RF}}(t) := \begin{cases}
%p_{\text{T}}(t) + h(\Delta p(t)), & \text{with probability } \text{prob}_{\text{RF}},\\
%\text{blank}, &   \text{otherwise}.
%\end{cases}
%\end{equation}

\subsection{HITL problem statement}

Define a binary absorbing state $x^{\text{ab}}_t \in \{0,1\}$, $x^{\text{ab}}_1 = 1$, that indicates when a simulation is terminated (i.e. $x^{\text{ab}}_t = 0$) s.t.
\begin{equation}
x^{\text{ab}}_{t+1} = x^{\text{ab}}_{t}(1- (timeout(t) \lor success(t) \lor defeat(t))), 
\end{equation}
where $timeout(t) \in \{0,1\}$ is a binary variable determining when time $t$ exceeds a pre-defined $timelimit \in \mathbb{N}$, i.e.,
\begin{equation}
timeout(t) =  \mathds{1}(t >time limit),
\end{equation}
$sucess(t) \in \{0,1\}$ is a binary flag indicating when ally drones win, i.e., 
\begin{equation}
sucess(t) = \mathds{1}(\sum_{i=1}^{n_{\text{ED}}} f^{i,\text{ED}}_t = 0) \prod_{i=1}^{n_{\text{ED}}} \mathds{1}(\|p^{i,\text{ED}}_t - p_{\text{RA}}\|_2 > r_{\text{RA}}),
\end{equation}
and $defeat(t)$ is a binary variable determining when enemy drones win, i.e.,
\begin{equation}
defeat(t) = \lor_{i=1}^{\mathbb{N}_{ n_{\text{ED}}}} \mathds{1}(\|p^{i,\text{ED}}_t - p_{\text{RA}}\|_2 \leq r_{\text{RA}}).
\end{equation}
In addition, for every ally drone $i \in \mathbb{N}_{n_{\text{AD}}}$ and enemy drone $j \in \mathbb{N}_{ n_{\text{ED}}}$, define the following relative positions
\begin{equation}
\Delta p^{i,j}_t := p^i_t - p^{j,\text{ED}}_t.
\end{equation}
Let $\mathbf x_t$  denote the global state at time $t$  including the above relative positions as well as those local states described in Sections~\ref{Sec:ally_state} and~\ref{Sec:enemy_state} as follows:
\begin{align}
\mathbf x_t &:= \{\Delta p^{i,j}_t | \forall (i,j) \in \mathbb{N}_{n_{\text{AD}}} \times \mathbb{N}_{n_{\text{ED}}}\} \cup x^{\text{ab}}_t \nonumber \\
 &\qquad \cup \{ \phi^i_t,f^i_t,g^i_t, p^{i,\text{GC}}_t | \forall i \in \mathbb{N}_{n_{\text{AD}}}\} \nonumber  \\
&\qquad \cup
\{\ell^{j,\text{ED}}_t, f^{j,\text{ED}}_t, g^{j,\text{ED}}_t, p^{j,\text{EDGC}}_t | \forall j \in \mathbb{N}_{ n_{\text{ED}}} \} .
\end{align}
Furthermore, let global action $\mathbf u_t$ be described as:
\begin{equation}
\mathbf u_t \hspace{-.1cm} := \hspace{-.1cm} \{ u^{i,\text{MA}}_t \hspace{-.1cm}, \hspace{-.05cm} u^{i,\text{SR}}_t \hspace{-.1cm}, \hspace{-.05cm}u^{i,\text{heading}}_t \hspace{-.1cm},\hspace{-.05cm} u^{i,\text{EMP}}_t \hspace{-.1cm}, \hspace{-.05cm}u^{i,\text{EJ}}_t \hspace{-.1cm}, \hspace{-.05cm} u^{i,\text{GPSS}}_t \hspace{-.1cm}, \hspace{-.05cm} u^{i,\text{EH}}_t | \forall  i \in \mathbb{N}_{ n_{\text{AD}}}\}.
\end{equation} 
We now define a reward function to take into account the task of tracking and neutralization of enemy drones. To this end, we define a negative reward function such that
\begin{equation}
\begin{aligned}
r_{\text{tracking}}(\mathbf x_t, \mathbf u_t)&:= - \sum_{i=1}^{n_{\text{AD}}} \sum_{j=1}^{n_{\text{ED}}} f^i_t f^{j,\text{ED}}_t \ell^{j,\text{ED}}_t \tanh(\|\Delta p^{i,j}_t\|_1) \\
& \quad -  \sum_{i=1}^{n_{\text{AD}}}  f^i_t \tanh(\|[u^{i,MA}_t, u^{i,SR}_t]\|_1).
\end{aligned}
\end{equation}
It is to be noted that tangent hyperbolic is a uniformly bounded and strictly increasing function, which is a desirable property to overcome normalization issues that often cause numerical instability. Moreover, we define a positive reward function for neutralizing an enemy drone
\begin{equation}
r_{\text{neutralization}}(\mathbf x_t, \mathbf u_t) \hspace{-.1cm}:=\hspace{-.1cm}  \sum_{i=1}^{n_{\text{AD}}} \sum_{j=1}^{n_{\text{ED}}}  f^i_t  u^i_t f^{j,\text{ED}}_t \ell^{j,\text{ED}}_t \mathds{1}(\|\Delta p^{i,j}_t\|_2  \hspace{-.1cm} \leq \hspace{-.1cm} r_{\text{N}}),
\end{equation} 
where $u^i_t$ is defined in~\eqref{eq:nut_action}. Hence, the global total discounted  reward function is given by:
\begin{equation}\label{eq:reward_UAV}
\Exp{\sum_{t=1}^\infty \gamma^{t-1} \big( x^{\text{ab}}_t (r_{\text{tracking}}(\mathbf x_t, \mathbf u_t) +   r_{\text{neutralization}}(\mathbf x_t, \mathbf u_t))\big)}.
\end{equation}

\indent \textit{HITL problem of UAV example:}
We are interested to develop a scalable multi-agent RL algorithm  to maximize~\eqref{eq:reward_UAV} with the help of human input.

\section{Proposed solution for UAV example}\label{sec:UAV_sol}
In this section, we propose a hierarchical  scalable decentralized multi-agent HITL reinforcement learning algorithm with three layers: supervision, tactics and drone. 

\subsection{HITL reinforcement learning}
The job of the first layer (i.e., supervision) is to have a human supervisor overriding lower layers' decisions when needed, the second layer (i.e., tactics) is to assign an enemy drone to an ally drone, and the third layer (i.e., drone) is to track and neutralize its assigned enemy drone.

 For simplicity, we use the tactics of \emph{closest enemy}  in the second layer and focus attention on the third layer. In particular, the state and action spaces  of the third layer are independent of the number of drones (i.e., scalable), where state space is the relative position between an ally drone and its assigned enemy drone and action space is its speed ratio and moving angle. It is to be noted that the information structure is decentralized because  every ally drone does not require (centralized) information about other ally and enemy drones. This feature is highly desirable in large-scale systems, as it leads to a more efficient setup computationally compared to its centralized counterpart. In addition, it is economically and physically more feasible. Moreover, it provides more privacy and robustness in the case when a drone is hacked or has faulty information.

Following the approach proposed in Section~\ref{sec: proposed_sol}, we use \emph{behavior cloning} as a special case of imitation-learning algorithm for the actor network to take advantage of the human advice. In addition, for the purpose of self-learning, we perform a pure actor-critic algorithm in the case when no advice is available. Finally, we build a critic network to gain some insights and intuitions about the solution. It is to be noted that one could directly find an actor (policy) without building a critic network (i.e. Q-function). The advantage of having Q-function is to build some intuitions about the solution, which is useful for transfer learning.
 
We consider two types of HITL: I) training phase and II) testing phase. We also consider two types of experts: I) pseudo-human and II) human. For now, we use two pseudo-human models (an analytical/heuristic algorithm and a trained agent) in the training phase and one human expert in the testing phase, as a proof of concept. In future work, we plan to replace the models by an actual human. Furthermore, we design a setup that allows for several simultaneous different types of advice interactions between the expert and AI. More precisely, the advice can be given a) off-line (pre-training) and/or  b) on-line (interactive) as well as in the form of c) action, and/or d) reward and/or e) demonstration.

\subsection{Experimental platform architecture using Cogment}
\begin{figure}[t!]
\center
\includegraphics[scale=.25,trim={0cm 0cm 0cm 0cm}, clip]{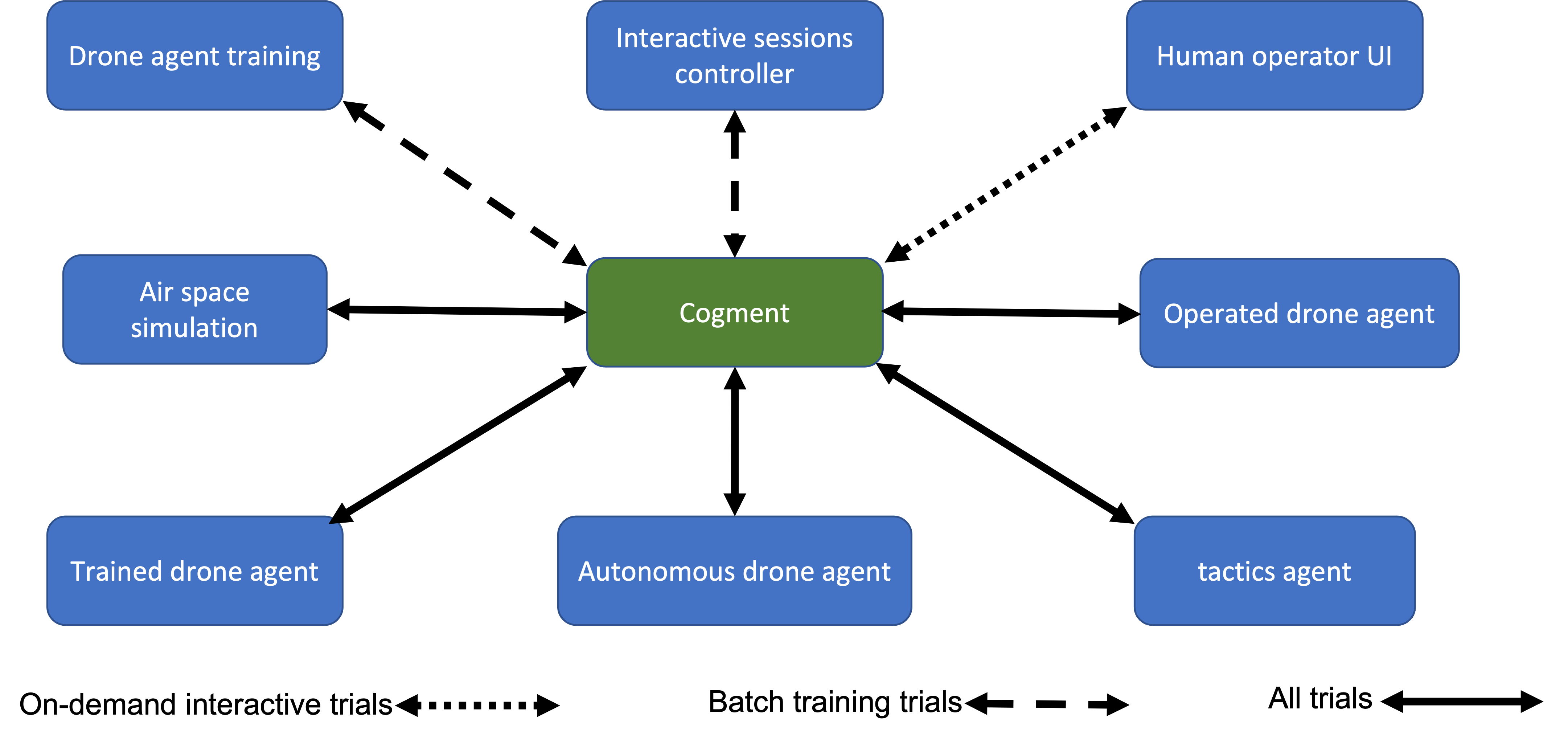}
\caption{Experimental platform architecture using Cogment.}\label{fig:Cogment}
\end{figure}

Figure~\ref{fig:Cogment} represents the architecture of the developed experimental platform. The Cogment platform~\cite{Cogment2021} is used to handle the orchestration of the execution and communication between the following different components:
%\begin{itemize}
%\item the agents, encapsulated in dedicated micro services as Cogment actors,
%\item the simulation, encapsulated in a dedicated micro service as a Cogment environment,
%\item the human operator UI, encapsulated as a client Cogment actor,
%\item the training process, as a python script relying on the Cogment SDK to trigger trials and retrieve activity data generated by them,
%\item the interactive session controller, as a component of the front-end relying on the Cogment SDK to trigger trials.
%\end{itemize}
(a) the agents, encapsulated in dedicated micro services as Cogment actors, (b) the simulation, encapsulated in a dedicated micro service as a Cogment environment, (c) the human operator UI, encapsulated as a client Cogment actor, (d) the training process, as a python script relying on the Cogment SDK to trigger trials and retrieve activity data generated by them, and (d) the interactive session controller, as a component of the front-end relying on the Cogment SDK to trigger trials.
Cogment dispatches observations of the environment from the simulation to the agents, as well as instructions from higher level agents. It then dispatches agents’ actions to the environment, which updates the simulation and agents instructions. Furthermore, a priority-ordered list of multiple agents can be assigned to a single drone entity at once. If a higher priority agent outputs a velocity/rotation change it overrides lower priority one thus enabling the human operator dynamic takeover. Cogment also provides its own trial data storage and make them available upon request.

\section{Numerical results}\label{sec:result}

We evaluate the performance of our algorithms based on \emph{success rate}. The success rate is defined as the number of times the ally team wins over the total number of experiments. To compute it, we train our model for $1000$ trials and  save the trained model every $10$ trials. Every trial starts with a random configuration of drones and lasts until whether one team wins or time is out. In this simulation, time-out limit is $2000$ and each trial often takes a few $100$ time steps to finish. We repeat the training process $10$ times, and save the corresponding trained models. Later, we test our saved models on various environments and compute the mean and standard deviation of success rate for each model over $10$ experiments. In Table~\ref{table}, numerical parameters are provided.

\begin{table}
\caption{Distances are in meters.}\label{table}
\centering
\resizebox{\columnwidth}{!}{%
\begin{tabular}{|c|c|c|c|}
\hline 
Parameter & Value & Parameter & Value \\ 
\hline 
$r_{\text{GC}}$ & $3000$ & $r_{\text{EGCS}}$ & $3000$ \\ 
\hline 
$r_{\text{N}}$ & $10$ & $\phi_{\text{EO}}^{\text{max}}$ & $\frac{300}{180} \pi$ \\ 
\hline 
$r_{\text{AD}}$ & $500$ & $v^{\text{max}}_{\text{SR}}$ & $10$ \\ 
\hline 
$r_{\text{RA}}$ & $100$ & $v^{\text{max}}_{\text{AS}}$ & $0.2$ \\ 
\hline 
$r_{\text{EO}}$ & $250$ & $v^{\text{max}}_{\text{EO}}$ & $\pi$ \\ 
\hline 
$r_{\text{RF}}$ & $2000$ & $v^{\text{max}}_{\text{GR}}$ & $0.2$ \\ 
\hline 
$r_{\text{GR}}$ & $3000$ & $pr_{\text{EMPD}}$ & $0.25$ \\ 
\hline 
$\rho_{\text{GR}}$ & $\frac{\pi}{2}$ & $pr_{\text{GR}}=pr_{\text{DR}}=pr_{\text{EO}}=pr_{\text{RF}}$ & $0.95$ \\ 
\hline 
$\rho$ & $\frac{\pi}{2}$ & $d_{\text{DE}}$ & $10$ \\ 
\hline 
$\rho_{\text{EO}}$ & $\frac{\pi}{3}$ & $d_{\text{AE}}$ & $\frac{\pi}{180}$ \\ 
\hline 
$p_{\text{GR}}$ & $(2000,3000)$ & $p_{\text{RA}}$ & $(3000,3000)$ \\ 
\hline 
\end{tabular} %
}
\end{table}

\begin{figure}[b!]
\center
\includegraphics[scale=.4]{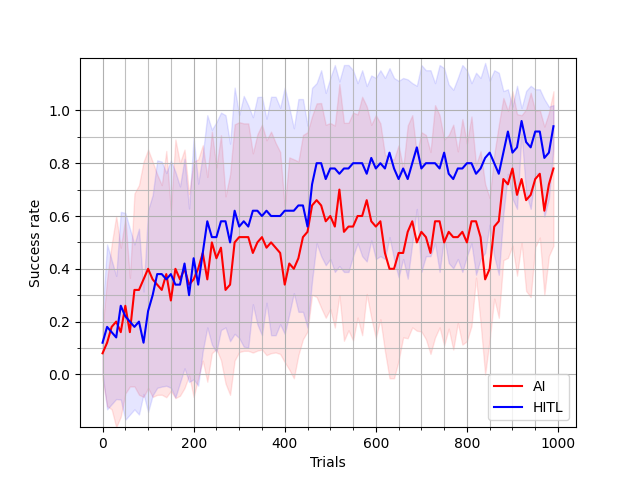}
\caption{HITL with 10 percent advice}\label{fig:HITL_10}
\end{figure}

\begin{figure}[b!]
\center
\includegraphics[scale=.4]{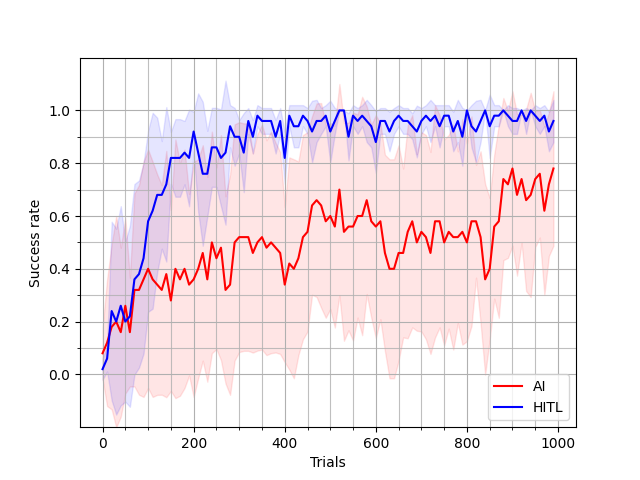}
\caption{HITL with 20 percent advice}\label{fig:HITL_20}
\end{figure}

Figure~\ref{fig:HITL_10} shows that HITL solution leads to better performance and less training time compared to AI solution. In this experiment, we interactively provide two types of advice: action and reward. More precisely, we use the expert’s advice $10\%$ of the time randomly during the learning process. In Figure~\ref{fig:HITL_20}, we increase the amount of advice from $10\%$ to $20\%$ and observe that the performance of HITL improves. Another interesting observation is that the variance of HITL solution decreases as the amount of advice increases from $10\%$ to $20\%$. This is due to the fact that advice provides a guiding direction for gradient methods so they less likely get stuck in local minima and more likely find the global point. 

In addition, we increased the amount of advice to $100\%$, which is a full imitation mode. In such a case, we observe that the HITL solution was sensitive to training configurations as well as the provided advice, and it performed poorly on the testing configurations. The reason for this behaviour is that AI was not given an opportunity to explore and learn the solution independently.

%
%We applied the learned strategy to the case in which the enemy's speed was 10 m/s and the success rate was 91\%. To compute the success rate, we calculated  the number of times red drones were neutralized over 100 trials.

\begin{figure}[t!]
\center
\includegraphics[scale=.4, trim={0cm 0cm 0cm 1cm}, clip]{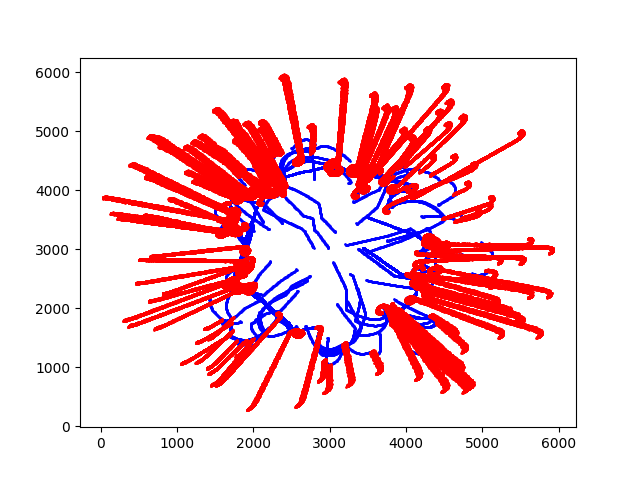}
\caption{Trajectories of $50$ ally drones in blue vs. $100$ enemy drones in red (i.e. overloaded attack).}\label{fig:HITL_50_50}
\end{figure}

To demonstrate the scalability of the proposed solution, we consider an \emph{overloaded attack}, where $50$ ally drones are met with $100$ enemy drones.  It is observed in Figure~\ref{fig:HITL_50_50} that the HITL DRL, trained with $20\%$ advice, is able to successfully neutralize the attack in a short period of time. This showcases the importance of human-AI collaboration in real-world problems because no human operator would be able to neutralize such a  large-scale attack in real-time.

\begin{figure}[b!]
\center
\includegraphics[scale=.4, trim={0cm 0cm 0cm 0.5cm},clip]{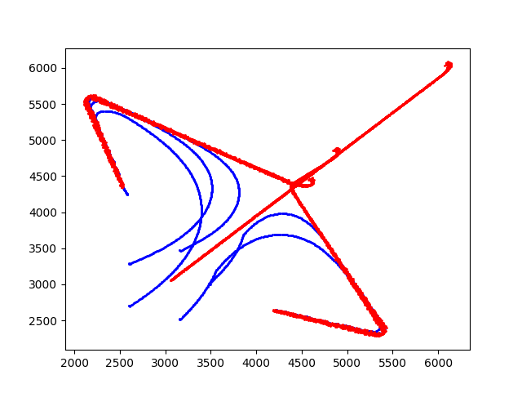}
\caption{Naive AI fails in decoy attack, where trajectories of ally drones are in blue and those of enemy drones in red. }\label{fig:AI_decoy}
\end{figure}

\begin{figure}[b!]
\center
\includegraphics[scale=.4,trim={0cm 0cm 0cm 0.5cm}, clip]{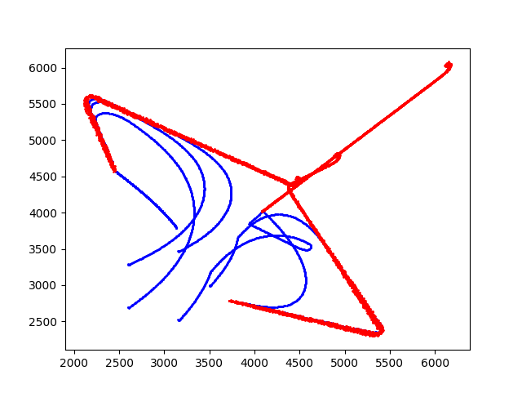}
\caption{HITL succeeds in decoy attack, where trajectories of ally drones are in blue and those of enemy drones in red. }\label{fig:HITL-decoy}
\end{figure}

In addition, we consider a \emph{decoy attack} with $5$ ally drones and $3$ enemy drones. In particular, the enemy drones attack the restricted area in such a way that the two closest enemies to the area (which seem to be more imminent threats to the naked eye) play as decoys to distract the attention from the actual threat (the third enemy). In Figure~\ref{fig:AI_decoy}, we observe how our previously trained AI is deceived by such a complex (unseen) attack and in Figure~\ref{fig:HITL-decoy}, we observe that how a human operator can cooperatively assist ally drones to neutralize the attack. This showcases the importance of human context awareness and experience in real-world scenarios.

%Next, we observe that our algorithm is able to learn  from human information more efficiently in simpler tasks than complex ones.

%good human  demonstration help us escape from getting stuck in  local minima (hence increasing performance and shortening convergence time). human advice can also save us a lot in efficient problem forumation of AI and hyper-parametrization. In general, having a well-formulated AI aith well-trained parameters as well as hyper-parameter can be very costly in both time and personnel. We can achieve a similar expertise by a decent AI and human. bad human demonstration will mislead resulting in worse performance and longer training time.

\section{Conclusions}\label{sec:conclusion}

Human-in-the-loop (HITL) deep reinforcement learning (DRL) represents a promising approach for improving the performance of RL agents in real-world applications.  To this end, 
a multi-layered hierarchical HITL DRL algorithm  was proposed. Three types of human inputs were taken into account: reward, action and demonstration. To verify the proposed solution, a simplified real-world UAV example was considered wherein it was shown that HITL resulted in faster training and higher performance. 
 As future work, we plan to consider more complex scenarios and collect more human inputs to  be able to observe more sophisticated and nuanced behaviours. In addition, we are interested to study situations in which human  can learn from AI.

\bibliographystyle{IEEEtran}
\bibliography{Jalal_Ref}
\end{document}